# Neuromorphic Online Clustering and Classification

J. E. Smith

University of Wisconsin-Madison (Emeritus)

October 26, 2023

**Abstract**

The bottom two layers of a neuromorphic architecture are designed and shown to be capable of online clustering and supervised classification. An active spiking dendrite model is used, and a single dendritic segment performs essentially the same function as a classic integrate-and-fire point neuron. A single dendrite is then composed of multiple segments and is capable of online clustering. Although this work focuses primarily on dendrite functionality, a multi-point neuron can be formed by combining multiple dendrites. To demonstrate its clustering capability, a dendrite is applied to spike sorting – an important component of brain-computer interface applications. Supervised online classification is implemented as a network composed of multiple dendrites and a simple voting mechanism. The dendrites operate independently and in parallel. The network learns in an online fashion and can adapt to macro-level changes in the input stream.

Achieving brain-like capabilities, efficiencies, and adaptability will require a significantly different approach than conventional deep networks that learn via compute-intensive back propagation. The model described herein may serve as the foundation for such an approach.

## 1. Introduction

The long term goal of neuromorphic computer architecture is to develop a large-scale computing device that can achieve brain-like abilities in a brain-like manner. Research toward such an ambitious goal will progress naturally as a sequence of steps with initial computing devices having cognitive abilities that fall far short of true thinking but can still perform useful functions efficiently and in a brain-like way. Two such functions are *clustering* and *classification* – functions at the root of modern-day machine learning and the topic of this document.

The approach taken is to reduce biological behavior to a plausible mathematical computing model (an architecture) that leads to simple, efficient implementations. The actual reduction process that leads to the proposed architecture is not covered in this document, only the resulting mathematical model. Briefly, it is a weighted synapse model based on active dendrites [15] that sum weighted synaptic inputs to produce outputs. Computing networks are constructed of active dendrites combined with other basic neural components.

An important aspect of brain-like behavior is online operation: each cycle, the computing model takes an input from an input stream, performs *inference* and *update* operations, and produces a new output as part of an ongoing output stream.

### 1.1 Clustering

With online *clustering*, a stream of input patterns (vectors) is applied, and a single dendrite groups them according to similarity. Each cluster of similar patterns is associated with a cluster identifier. As the inputs are streamed in, the network first infers the cluster to which the current input pattern best belongs and provides the cluster identifier as the output. Then the current input pattern is assimilated into the learned sequence of input patterns via an update operation that adjusts synaptic weights contained in the active dendrites. Inference is performed every cycle to produce an output. Weight updates can be turned on and off by a controlling agent.

## 1.2 Classification

*Classification* is performed by multiple dendrites that learn via online supervision. Generically speaking, input patterns are grouped into classes and each class has a *label*. There are as many dendrites as there are labels. When a streamed input pattern is applied, the network first infers (predicts) its label and provides it as the output. The correct label is then given as a supervisory input, and an update function adjusts synaptic weights in the clustering dendrites accordingly. As with clustering, update can be turned on and off by a controlling agent. In practice, the agent will typically turn on update for some initial sequence of input patterns (to warm up, or train, the weights), and then turn off training when labels are unavailable. When training is turned off, inference-only operation predicts the label for each input based on what it learned during supervised training, i.e., it classifies the input pattern. In general, training can be periodically turned on and off by the agent as the system runs.

## 1.3 Spiking Neural Networks

Spiking Neural Network (SNN) research spans more than three decades, and the set of proposed networks generally referred to as "SNNs" is huge and widely diverse. Just about the only thing they all have in common is that inter-neuron communication is based on "spikes", typically represented as binary 1s, with a non-spike being a binary 0. The networks studied in this document belong to a specific class of SNNs satisfying the following criteria.

1) The presence (or absence) of individual spikes conveys information, rather than the rate at which spikes are emitted.

2) The basic computational function is that of a *point neuron* (see Figure 1), although in this work this function is performed by a single dendritic *segment*.

3) Synaptic weights are updated via localized Hebbian learning rules.

4) Clustering is an essential building block function that is implemented by combining parallel point neurons (parallel segments in this work) with a winner-take-all (WTA) circuit.

5) Network operation is synchronized, with a single layer of segment/WTA functionality per synchronization cycle.

SNNs having various subsets of these features have been proposed and studied for decades. Thus, the network architecture proposed here is based on existing concepts (sometimes long existing) combined in new and different ways.

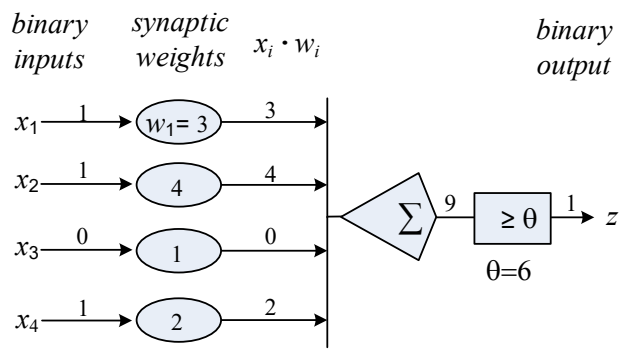

**Figure 1. A classical point neuron. The spikes (1s) in a binary input vector are multiplied by associated synaptic weights and then summed. If the sum (dot product) reaches a threshold θ, then the output is a 1; else it is 0. There are more complex point neuron models, but the key feature they have in common is a single summation point.**

## 1.4 Active Dendrites

Although the *functionality* of the classical point neuron is basic to the model, the functional element in the model is not referred to as a "neuron", rather it is a dendritic "segment". In this work, a multi-point neuron model is composed of multiple active dendrites [15] each composed of multiple segments. This multi-point neuron model is both more realistic and computationally more powerful than the classical point neuron. The difference between a point neuron and a multi-point neuron is analogous to the difference between an AND gate and a multiplexor.

## 1.5 Document Overview

Just as with conventional computer systems, a neuromorphic architecture can be organized as a hierarchical stack. Figure 2 shows the stack for a specific part of the brain: the neocortex. Other parts of the brain have architectures that share at least the lower levels of the stack, and this particular document is focused only on the lower levels. Higher levels (up to a single macrocolumn) are covered in another document [9].

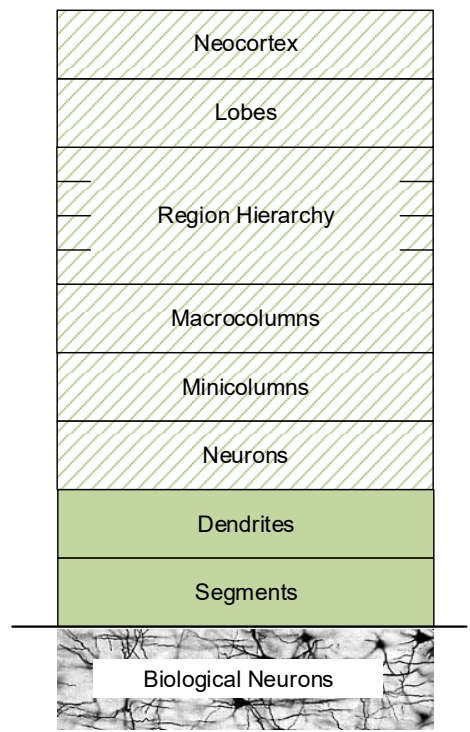

**Figure 2. Architecture stack for the neocortex. The lower layers, at least up to the neuron layer, are also part of the architecture for other parts of the brain. In this document only the segment and dendrite layers are covered in detail.**

Preliminaries regarding clustering are covered in Section 2. Then in Section 3 a single segment is shown to perform essentially the same function as a classic integrate-and-fire point neuron. In the active dendrite model (Section 4), a single dendrite is composed of multiple segments, each of which has multiple synaptic inputs. In Section 5, a single dendrite is shown to be capable of online clustering. A clustering network is developed and applied to spike sorting – an important component of brain-computer interface (BCI) applications. In Section 6, supervised online classification is implemented as a network composed of multiple, parallel dendrites and a simple tallying mechanism. The multiple dendrites operate independently and in parallel. The network learns in an online fashion and can adapt to macro-level changes in the input stream.

## 2. Preliminaries

### 2.1 Data Representation

Information is communicated as scalars, vectors, and 2-dimensional arrays, and there are two data types: *bits* and *ints* (integers).

*Bit* vectors are typically used at communication interfaces, and *int* vectors are used for certain computations intrinsic to the model. *Bit vectors* model neural spike volleys where a "1" indicates the presence of a spike. In the model, bit vectors encode information in two ways.

1) To specify sets of features:
    1 → feature is present.
    0 → feature is not present.
    Example: [01011] → features 2, 4, and 5 are present

2) Unary encoded values:
    The value being encoded is the 1's count.
    Example: [01101], [11100],[00111] all encode "3"

### 2.2 Clustering

Clustering partitions a set of patterns into groups – *clusters*. When forming clusters the objective is that the members of a cluster are more similar to each other than members of other clusters. There are many clustering methods based on different definitions of "similarity".

For our purposes, clustering begins with an ordered multi-set of row vectors (patterns), $P = [x^1, x^2, ..., x^{|P|}]$, where a given vector may appear one or more times in P. *Clusters* are disjoint subsets of P. The members of P are ordered because we are implementing online clustering methods where the order of applied inputs is important.

The *centroid* c of cluster C is a row vector consisting of component-wise averages over the members of the cluster. In this work, when centroids are computed directly (for evaluating metrics), the arithmetic mean is used. To be more precise, say cluster C is composed of $p$ row vectors each of size $q$. These can be represented as a $p \times q$ array $x^{1:p}_{1:q}$ Then the centroid $c$ is defined as a row vector where $c_j = (\Sigma_{i=1:p} x^i_j)/p$.

Of interest here are clustering methods that take an input stream of patterns (encoded as *bit* vectors) and compute a set of centroids via an online *learning* process. Given a set of centroids, an *inference* process computes distances between a given input pattern and each of the centroids, thereby identifying the nearest centroid, i.e. the one most similar to patterns belonging to the associated cluster. The output of this inference process is a *cluster identifier* (CId). With *online* clustering, the training process interleaved with inference. Consequently, if input similarities shift over time, the centroids, and therefore the clusters, also shift to reflect the changes.

In this document, the sum of absolute differences (*sad*) is used for computing distances. Given two row vectors $x$ and $y$, each having $q$ components, $sad(x,y) = \Sigma_{i=1:q} | x_i - y_i |$; this is a rectilinear (Manhattan) distance metric. Given a set of $p$ clusters $C_{1:p}$ and their associated centroids $c^{1:p}$, the *nearest centroid* for vector $x$ is $c^i$ such that $sad(x, c^i) \leq sad(x, c^j)$ for all $i \neq j$.

An *average distance* metric for a given set of clusters and associated centroids is found by averaging the distances between the patterns and their associated centroids.

As a basis for performance comparisons, the *k-means* algorithm [20] is used. It is an offline epoch-based compute-intensive process that begins with a set of $k$ randomly or heuristically placed centroids and repeats 2-step epochs: 1) determine a new set of clusters based on nearest centroids, 2) determine a new set of centroids for these clusters. These two steps repeat until the centroids converge. For a given set of clusters and associated centroids, the *k-means convergence* metric is the fraction of patterns that are nearest to the centroid of their currently assigned cluster. The *k-means* algorithm, as usually described, converges to 1.0. However, achieving a convergence value slightly less than 1 (say .98) is adequate in many situations and is computationally more expedient.

## 2.3 Nearest Centroid Computation

In this work, the components of a centroid are approximated by synaptic weights, and the inference process determines the nearest centroid based on the synaptic weights. An essential part of the method is that bit vectors being clustered generally have a constant weight[1], i.e., they all have the same number of ones. However, as a practical matter, some small variations from the constant weight may occur. Assuming constant weight vectors simplifies the determination of the nearest centroid, especially if the vectors are also sparse. The mathematical basis for this simplification is outlined in this section.

Let cluster C be composed of $p$ row vectors each of size $q$. These can be represented as a $p \times q$ array $x^{1:p}_{1:q}$. Then the centroid $c$ is defined as a row vector where $c_j = (\Sigma_{i=1:p} x^i_j)/p$. Summing all the components of a centroid yields

$$\Sigma_{j=1:q} c_j = (\Sigma_{j=1:q} \Sigma_{i=1:p} x^i_j)/p.$$

The summations can be interchanged so

$$\Sigma_{j=1:q} c_j = (\Sigma_{i=1:p} \Sigma_{j=1:q} x^i_j)/p.$$

If each row vector has exactly $m$ ones, then $\Sigma_{j=1:q} x^i_j = m$ and summing over all row vectors yields

$$\Sigma_{i=1:p} \Sigma_{j=1:q} x^i_j = mp.$$

Consequently, using substitution

$$\Sigma_{j=1:q} c_j = mp/p = m.$$

(From here on, all summations are over $j=1:q$). Given bit vector $x$,

$$\Sigma c_j = \Sigma x_j c_j + \Sigma \sim x_j c_j = m.$$

Break *sad* $(x, c)$ into two parts – one with spikes ($x_j=1$) and one without ($\sim x_j = 1$):

$$sad(x, c) = \Sigma x_j |x_j - c_j| + \Sigma \sim x_j |x_j - c_j|$$

$$\Sigma x_j |x_j - c_j| = \Sigma x_j - \Sigma x_j c_j = m - \Sigma x_j c_j$$

$$\Sigma \sim x_j |x_j - c_j| = \Sigma \sim x_j c_j = m - \Sigma x_j c_j \quad \text{because } m = \Sigma x_j c_j + \Sigma \sim x_j c_j$$

Then

$$sad(x, c) = 2(m - \Sigma x_j c_j).$$

Consequently, the *sad* is minimized when $\Sigma x_j c_j$ is maximized. This means that to determine the nearest centroid, one need only compute $\Sigma x_j c_j$ for each centroid because the centroid that yields the maximum $\Sigma x_j c_j$ is the nearest centroid.

## 3. Segments

The *segment* is a fundamental computational unit (Figure 3). The schematic notation is described in Appendix 1. A segment is akin to a classic spiking point neuron except the output is not a spike determined by a threshold value. Rather, the output is the result of the dot product. A threshold value ($\theta$) is applied, nevertheless, so if the dot product is less than the threshold the output value is 0.

---

[1] To resolve ambiguity: this usage of "weight" is different from the synaptic "weight".

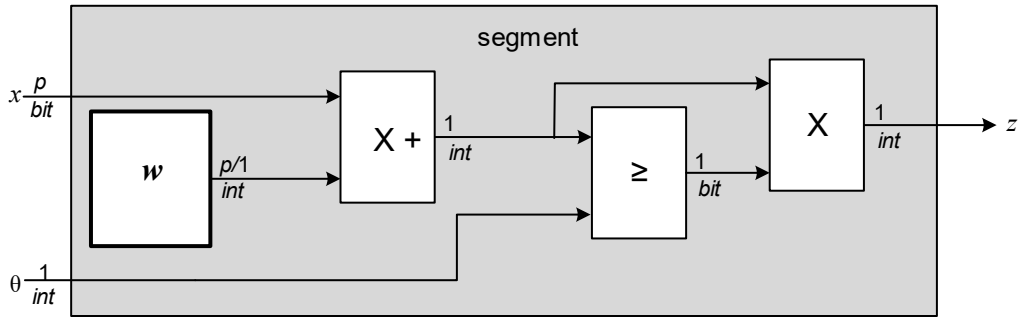

**Figure 3. Schematic for a segment.** Input *x* is communicated as a *p bit* row vector. Weights, *w*, are expressed as a *p int* column vector. When a *bit* vector is applied to the synapses, the product is an *int* scalar that is the body potential for the given segment. If the potential reaches the threshold value θ, the value of the potential is passed through to the output *z*. Observe that when multiplication takes place, one of the operands is always a *bit* or a *bit* vector, thereby simplifying the "multiplication" significantly.

## 4. Dendrites

*Dendrites* perform an (approximate) clustering function. The method is simple and amenable to online, localized learning. Before defining dendrite function precisely, a less formal illustration of clustering performed by a trained dendrite is in Figure 4.

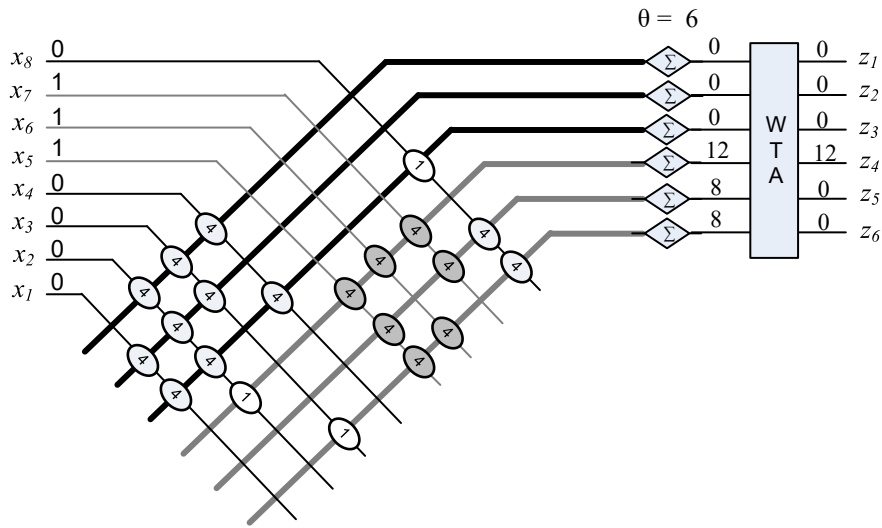

**Figure 4. An illustration of dendrite clustering.** Inputs from bit vector *x* are applied to a synaptic crossbar. The maximum synaptic weight is 4, and synapses with 0 weight are not shown to reduce clutter. A bimodal weight distribution is typical. Each of the six segments performs a dot product *x · weights* and produces an *int* output value (this value is the segment's *potential*). Then, provided the potential reaches the threshold of θ = 6, the maximum potential (12 in this case) is passed through to winner-take-all (WTA) output $z_4$ ; all the other components of vector *z* are 0.

### 4.1.1 Winner Take All (WTA) Inhibition

The schematic for WTA inhibition is in Figure 5. If there is a tie for the maximum value, and with low precision integers ties will be relatively common, then the tie must be broken by choosing one. In the model simulated later, the tying input with the lowest index (subscript) is selected. This is a systematic method that makes simulator debugging easier. However, in general the "winner" of a tie can be arbitrarily chosen, including pseudo-random selection.

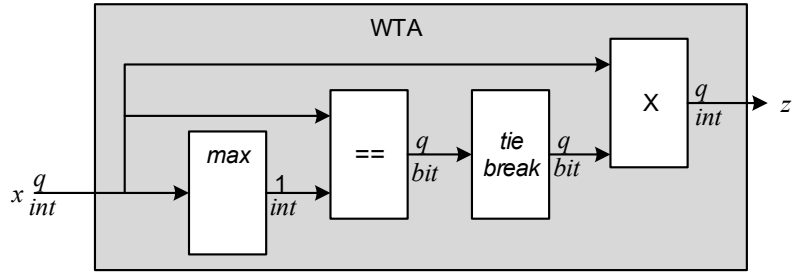

**Figure 5. Schematic for WTA inhibition. First, the maximum value of a size *q int* vector *x* is found. Then the "= =" block selects the input positions that have the maximum value. If there is more than one (a tie), then a *tie break* mechanism selects one. The output vector *z* contains 0s in every component except the component with the maximum *x* value, in that component *z* is assigned the value in *x*.**

In general, the top *n* values can be selected via *n*-WTA. In this document, 1-WTA is commonly used, so unless stated otherwise, the standalone "WTA" implies 1-WTA.

### 4.1.2 Active Dendrites

The schematic for an active dendrite is given in Figure 6. It combines parallel segments with WTA and includes an enable signal. Given a set of synaptic weights, a dendrite partitions input patterns into clusters. In the active dendrite model as used here, the input patterns to be clustered are applied to *distal* inputs (D in the schematic) and the *proximal* input (P) acts as an enable.

All segments receive the same input, *x*, via a synaptic crossbar. *x* is a *p* component row vector and $w^{1:p}{}_{1:q}$ is a $p \times q$ weight matrix. The vector *x* is multiplied by the weight matrix *w*. The product, a *q* component vector, passes through WTA inhibition. Only the maximum component of the product is passed through to the WTA outputs (as long as it satisfies the threshold θ).

The output vector *z* is coded as a 1-hot $int^2$. The position of the 1-hot value indicates the cluster to which the input pattern belongs, i.e. the CId, and the value indicates how close the input is to the centroid: a higher potential indicates the input is closer to the centroid.

In this model, the weights are an approximation of the centroids (multiplied by $w_{max}$). This will be illustrated later (Figure 17). As shown in Section 2.3, finding the nearest centroid for input *x* is equivalent to finding the centroid with the maximum, $\Sigma x_j c_j$. Using the weights as approximate centroids, the dendrite computes the maximum $\Sigma x_j w_j$, thereby yielding the nearest approximate centroid.

---

[2] The term "one-hot" usually refers to a single 1 *bit*. However, in this work the concept is extended so a vector with a single non-zero *int* is also "one-hot".

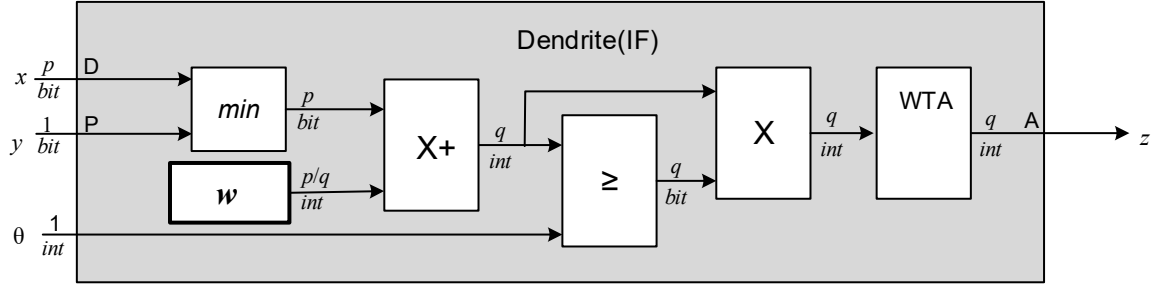

**Figure 6.** Schematic for the dendrite inference function (IF) in detail. Update function (UP) detail is in Figure 9. An active dendrite contains *q* segments that operate on distal inputs (D). A 1-bit proximal input (P) acts as an *enable* signal, implemented as a *min* operation. The result is multiplied by a *p* × *q* synaptic weight matrix. The resulting *q int* vector is checked against the threshold θ, producing a *q bit* vector. This *bit* vector when multiplied by the *int* vector selects all the values reaching the threshold. WTA inhibition finishes the job by passing through the largest of the values to output A; all other WTA outputs are 0. Note that for the update function (Figure 8), the *int* output A is binarized to *bits*, i.e., spikes.

Although one could potentially model more than one proximal input per dendrite, in the model developed thus far there is only one.

As one interpretation, the distal input spikes define a specific *context*, and a proximal spike indicates a specific *object* or *feature* that is present at the specified context. In another interpretation, a distal input defines a *pattern*, and a proximal spike indicates an associated *label*. Often the distal input vector is the concatenation of multiple vectors, each being associated with some aspect of the context.

In summary: each of the segments is associated with an approximate cluster centroid, the synaptic weights approximate the centroid values, and the inference process determines the CId of the nearest approximate centroid.

*Online Learning*

Spike dependent plasticity (SDP) is a binarized version of the classical Spike Timing Dependent Plasticity (STDP)[10][13]. See Table 1.

**Table 1. SDP update functions.**

| $x_i$ | $z_j$ | weight update | limitation |
|---|---|---|---|
| 0 | 0 | $\Delta w^i_j = 0$ | |
| 0 | 1 | $\Delta w^i_j = -backoff$ | $w^i_j \geq 0$ |
| 1 | 0 | $\Delta w^i_j = +search$ | $w^i_j \leq w_0$ |
| 1 | 1 | $\Delta w^i_j = +capture$ | $w^i_j \leq w_{max}$ |

SDP is performed by each synapse and updates the synaptic weight by an amount that depends on the presence or absence of a synapse's input spike and the output spike of the synapse's associated dendrite (after it has been binarized). The SDP update function for weight $w^i_j$ in Table 1 takes bit $x_i$ as one input and bit $z_j$ as the other. The output is an *int* $\Delta w^i_j$ indicating an amount by which $w^i_j$ should be updated.

For *backoff* and *capture* updates, weights saturate at 0 and $w_{max}$. For the *search* update, there is a lower saturating weight $w_0$.

*Cluster Formation*

The SDP parameters *backoff*, *capture,* and *search* control the formation and evolution of clusters. In particular, the relative values of *capture* and *backoff* affect the distance (or alternatively *overlap*) between members of the same cluster. It is unlikely the optimal relationship can be reduced to a simple mathematical expression that holds for realistic data, so *capture* and *backoff* are determined via offline parameter sweeps using data that is similar to the data to be observed in practice.

Nevertheless, it is useful to consider parameter relationships in a qualitative way by studying individual cases. For example, once a cluster is established and a new input vector is applied, whether the new vector should become part of the existing cluster or should begin a new cluster is related to the *overlap* between the vectors. If a high amount of overlap (low distance) is desired then the *capture* and *backoff* parameters should be biased toward starting a new cluster. If a low amount of overlap is sufficient, then the parameters should be biased toward joining the existing cluster.

To illustrate this, consider the situation where all the input bit vectors contain exactly *m* spikes. Define the *overlap* to be the number of bits that two vectors have in common. Assume *capture* is fixed at 1 and derive a relationship between *overlap* and *backoff*. Say two vectors with a given *overlap* are consecutively applied to an un-trained system with weights initialized to a baseline value $w_0$. As shown in [9] they will map to the same cluster if

    *backoff* < *overlap*/(*m*-*overlap*).

Otherwise, they will be placed in different clusters. See the example in Figure 7. Qualitatively, increasing *backoff* will tend to increase the amount of overlap required for two vectors to be placed in the same cluster: higher *backoffs* will tend to create more clusters.

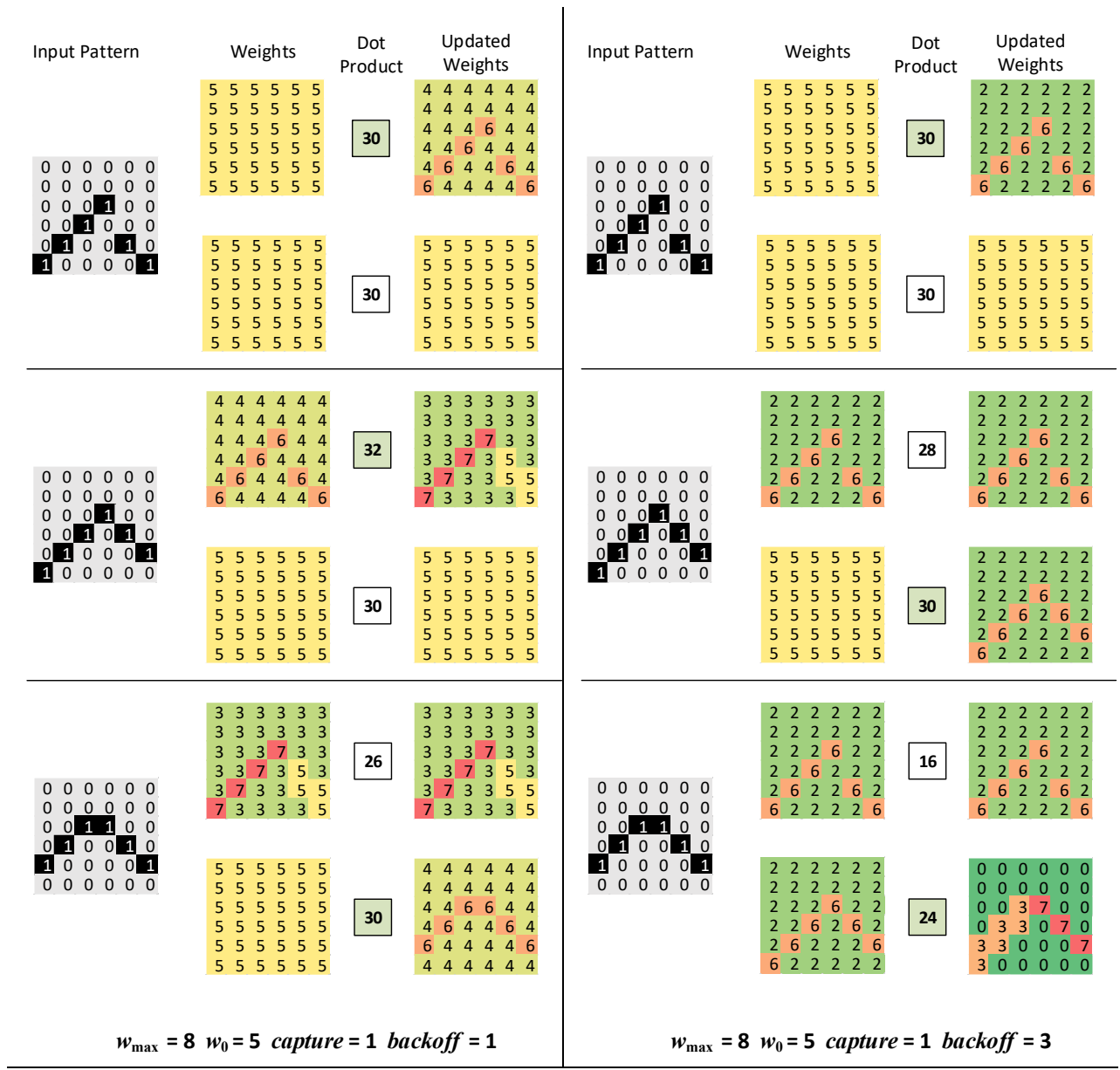

Figure 7. Two examples of the way *capture* and *backoff* influence cluster formation. In both the left and right columns, three consecutive patterns are applied, each having 6 bits set to 1 ($m = 6$). If it is desired that two patterns go into the same cluster when the *overlap* is 4 or more, then *backoff* < $4/(6-4) = 2$. In the left column, backoff = 1, so overlap of 4 or more will map patterns to the same cluster. Three patterns are applied along the leftmost column. There are two clusters with synaptic weights initialized to $w_0 = 5$. When the first pattern is applied (upper left), the dot products for both clusters are 30, a tie. The first cluster is arbitrarily selected to fire an output spike and update its synaptic weights. The updated weights are shown to the right of the dot products. The next input pattern overlaps with the first in 4 bit positions, so it should go into the same cluster. The dot products are 32 and 30, so the first cluster is selected by WTA. The third pattern has an overlap of less than 4 with respect to the previous 2. Consequently, the dot products are 26 and 30, and the third pattern goes into a new cluster. For the example in the right column, backoff = 3; consequently an overlap > 4 will be required if the first and second patterns are to map to the same cluster. In the example, the overlap is not >4, so the two patterns map into different clusters.

*Search*

The *search* parameter allows clusters to adapt to long-term input changes. If a segment $z_i$ has captured a cluster and the input stream changes so that the cluster no longer reflects patterns in the changed input stream, $z_i$ may no longer spike. Whenever a synapse receives an input spike but there is no output spike, the synaptic weight is incremented by *search* (up to some maximum weight $w_0$, where $w_0 < w_{max}$). In practice *search* << *backoff* and *capture*. Consequently, as long as $z_i$ does not spike, the synaptic weights for $z_i$ will gradually creep upward according to the input spikes they receive. Eventually, the weights will become high enough that output $z_i$ spikes and a new cluster is captured. Or, if a member of the original cluster appears, $z_i$ will spike, and *backoff* will decrease the weights of the searching synapses by an amount significantly greater than the *search* increment.

*Update Schematics*

Schematics for the *capture* and *backoff* update networks are in Figure 8. The weight matrix contains $p$ rows and $q$ columns ($p/q$). For *capture* (Figure 8a), there is a matrix multiplication of input $x$ transposed to a $p$ bit column vector and the $q$ bit row vector $z$, yielding a $p/q$ bit mask. The *capture* value is multiplied by the bit mask, and the product is added to the weights. A *min* function limits the weight to $w_{max}$. The *backoff* network (Figure 8b) works in a similar way, except the $x$ input is complemented ("~") in order to implement the update function given in Table 1. Similarly, *search* (not shown) complements the $z$ input.

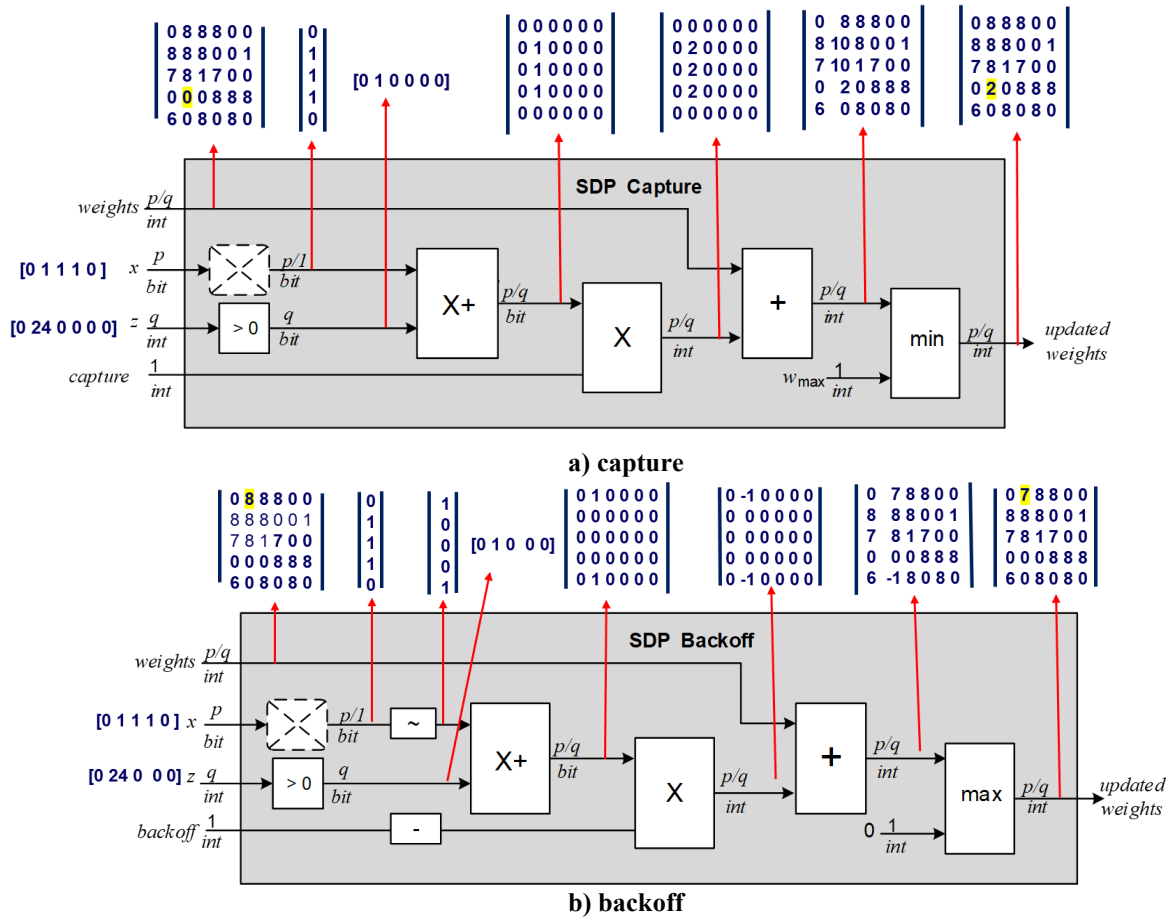

Figure 8. Schematics for SDP *capture* and *backoff* updates. Example vectors and arrays for a typical update are shown.

In the example just given, functions are defined for dense arrays and vectors for illustrative purposes. In most realistic applications, data will be very sparse (one-hot coding is ubiquitous). Hence, an efficient implementation will likely maintain data in a sparse (i.e., compressed) form.

The three update functions are mutually exclusive so they can be performed in parallel with their outputs merged before being added to the *weight* array (Figure 9).

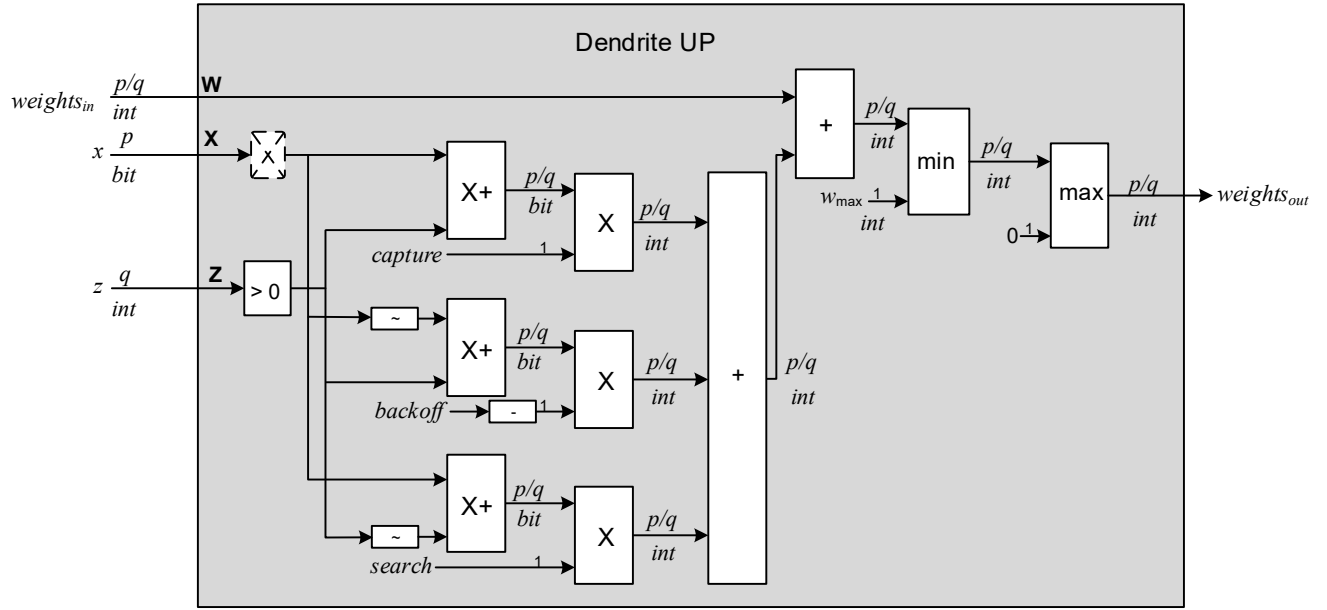

**Figure 9. Complete synaptic update network for a dendrite. The capture, backoff and search update values are combined prior to final update.**

# 5. Online Clustering with Active Dendrites

An active dendrite forms clusters controlled by an enable signal. If a single dendrite is used with the enable signal always active, then a single dendrite performs basic online clustering. This functionality is demonstrated via an example application: spike sorting.

## *5.1 Research Benchmark: Spike Sorting*

*Collaborators: Abhishek Bhattacharjee BCI group at Yale.*

Spike sorting is a key component of certain Brain Computer Interface (BCI) applications, as well as other applications where individual neuron behavior is of interest. An important BCI application is prediction of an on-coming brain seizure by detecting anomalous spiking behavior of individual neurons. As part of a BCI, tiny electrocorticographic (ECoG) probes are embedded into the brain (Figure 10). Typically these probes consist of multiple channels, but for the research benchmark used here, the activity from only one channel is considered (Figure 11).

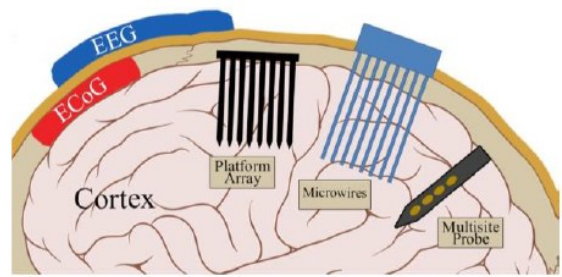

**Figure 10. Input for the spike sorting comes from electrical probes inserted into the brain. Of primary interest here are signals from ECoG probes placed under the skull.**

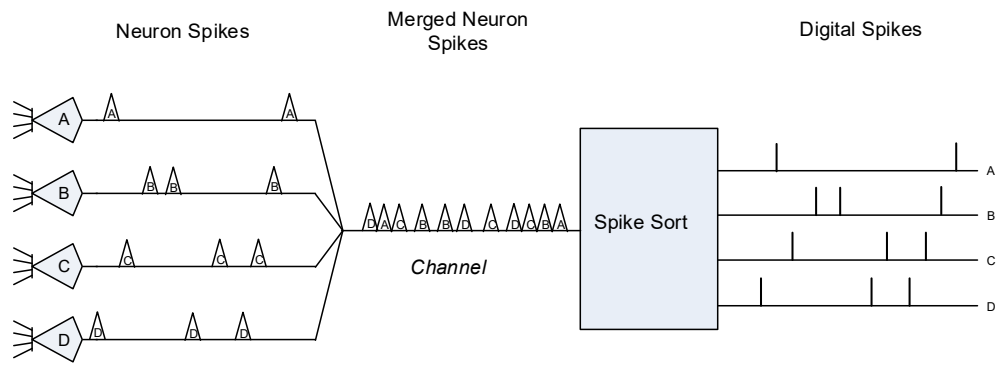

**Figure 11. A channel detects electrical activity from several nearby neurons (A,B,C,D), and the task of spike sorting is to separate and identify the activity of individual neurons. This is done by clustering action potentials (spikes) based on the shapes of their waveforms.**

The processing chain is shown in Figure 12. There is significant front-end processing that precedes the clustering application. The preprocessing step consists of filtering and spike detection. The filtered signal is used as input for encoding and clustering. Because encoding greatly affects the quality of clustering, it is included as part of the simulated implementation.

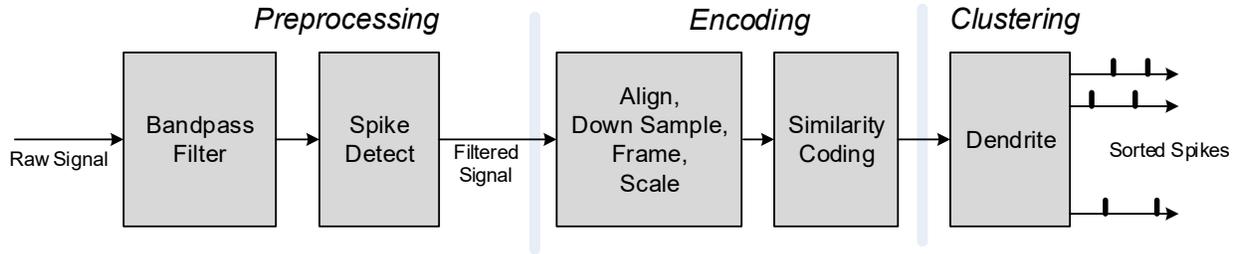

**Figure 12. Spike sorting.** Preprocessing consists of a bandpass filter and a spike detection mechanism. As part of encoding, the filtered signal is aligned to a spike's peak, and the signal is downsampled, framed, and scaled. The image is then similarity encoded (explained below). Finally, the preprocessed images are passed to the clustering function which is implemented as a single dendrite that is always enabled.

### 5.2  Spike Sorting Design Method

Using a good encoding method is essential for overall performance. Consequently, the design method used here is a two-step process. First, frontend encoding is optimized using a *k*-means backend for clustering. Although it is not an online method, using a *k*-means backend during frontend optimization essentially isolates the clustering process from the encoding process. After frontend encoding is optimized, the online clustering method is then optimized, yielding a complete, optimized, online design.

### 5.3  Encoder Optimization

Referring to Figure 12, it is assumed that pre-processing (filtering and spike detection) has been performed. The input to the encoding process consists of discrete time samples of voltage levels. That is, each input waveform is a 1D vector. The following parameters define the initial *framing* step.

*precision*: bits of input signal precision

*before* & *after*: the range of samples before and after the spike peak

*stride* & *window*: temporal parameters for block downsampling

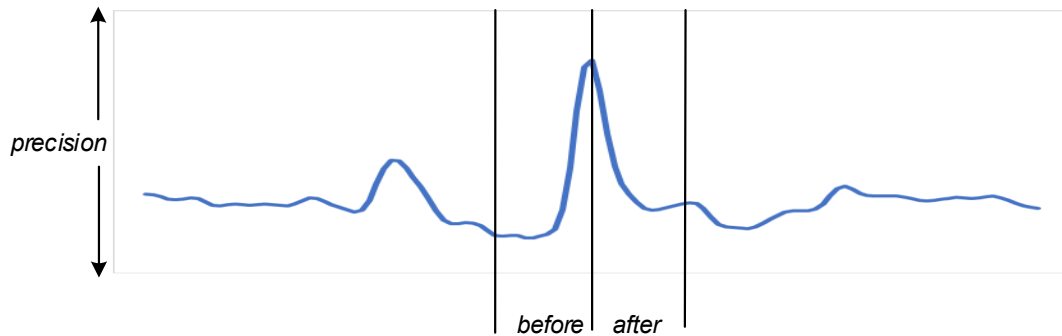

**Figure 13.** After filtering, the input signal is reduced to a given *precision* in the *y* dimension and framed to given times *before* and *after* the spike peak in the *x* dimension. The signal is also block downsampled in the *x* dimension using a *stride* and *window* size as parameters (not shown).

Given the above parameters, the waveforms are *scaled* by taking the maximum and minimum voltage levels over the entire benchmark and scaling the waveforms linearly so that at least one waveform reaches the maximum and one reaches the minimum. After framing and scaling, the 1D vectors are converted to a bit (pixel) representation. Conceptually, this is a 2D bit image although it is applied to the clustering unit as a long 1D *bit* array in row major format. The final step, *similarity coding*, is described in the next subsection. These steps are illustrated in Figure 14.

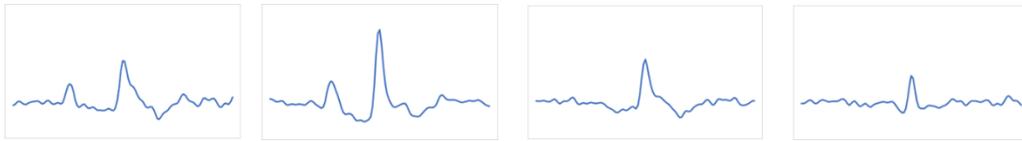

a) Typical waveforms after filtering and spike detection

[6 6 6 6 10 17 20 16 13 12 10 9 7 7]   [4 3 4 5 15 27 25 15 10 8 8 8 8 8]   [7 7 7 7 12 19 19 14 11 11 10 9 8 7]   [9 8 7 6 8 14 15 10 7 7 8 8 8 7]

b) Integer vectors after alignment, downsampling, framing, and scaling.

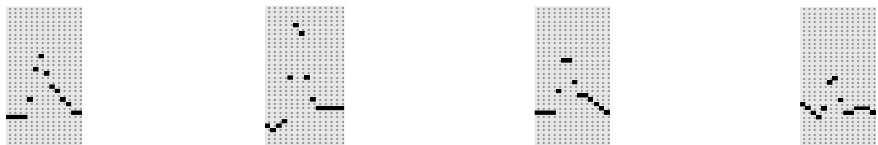

c) After conversion to bits (pixels).

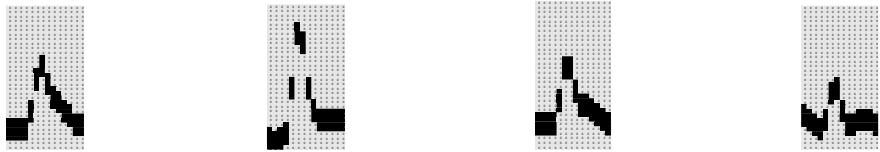

d) After similarity encoding

Figure 14. Examples of frontend processing steps.

## 5.4 Similarity Coding

When forming clusters, similarity is based on distance, and when bit vectors are used (as is the case here), the *sad* distance metric is Hamming distance.

One way of encoding values is to use a 1-hot code; the value *n* is represented as the 1-hot vector where the *n*th bit position is a 1 and all the others are 0. The value 3 is represented as [0 0 1 0 0 0]. Index values are naturally used in simulator code and are used for resolving ties in WTA inhibition to simplify simulator debugging, but index values are not available to the model as operands. Consequently, similarity of values cannot be determined by considering their indices, so 1-hot coding does not provide useful similarity information. See Figure 15a where values 3, 2, 5, and 6 are encoded 1-hot. The distance between the value 3 and all the others is 2, regardless of how close the value being encoded is to 3.

With similarity coding, the 1-hot code is converted to an *m*-hot code (where *m* is typically odd) by setting ⌊ *m/2*⌋ positions on either side of the encoded value to 1s. Consequently, the *sad* of *m*-hot code vectors provides a rough measure of distance (Figure 15b). In the example, 2 and 3 are distance 2; 5 and 3 are distance 4; and 3 and 6 are distance 6. Beyond a certain range, depending on *m*, all the values become the same distance. See [26] for a more formal discussion of similarity coding (although not by that name).

| value | encoding | distance from 3: |
|---|---|---|
| 3: | 0 0 1 0 0 0 0 | |
| 2: | 0 1 0 0 0 0 0 | 2 |
| 5: | 0 0 0 0 1 0 0 | 2 |
| 6: | 0 0 0 0 0 1 0 | 2 |

a) 1-hot

| value | encoding | distance from 3: |
|---|---|---|
| 3: | 0 1 1 1 0 0 0 | |
| 2: | 1 1 1 0 0 0 0 | 2 |
| 5: | 0 0 0 1 1 1 0 | 4 |
| 6: | 0 0 0 0 1 1 1 | 6 |

b) 3-hot

**Figure 15. Examples of 1-hot and 3-hot similarity coding. With 1-hot coding encoding for all other values are the same distance from the encoding for 3. With 3-hot encoding, the distances from other encodings reflect differences in the encoded values.**

When *similarity coding* is applied to the 2D spike patterns (e.g. Figure 14c) , the parameters are:
   *vertical hotness* : the number of similarity bit positions ($m$) in the y-dimension
   *horizontal hotness*: the number of similarity bit positions ($m$) in the x-dimension

When similarity coding with vertical hotness of 3 and horizontal hotness of 1 is applied to the example in Figure 14c, the images in Figure 14d are the result.

### 5.5  Clustering

For clustering, all segments share the same global parameters: $w_{max}$, $w_0$, *threshold*, *capture*, *backoff*, and *search*. These are determined by systematically exploring the parameter space. This is done with an offline training data set having characteristics that are similar to the data used for testing. For the results reported here, however, the same dataset is used for both establishing global parameters and performance testing.

The general procedure is to first set values for $w_{max}$ and $w_0$, then set *capture* = 1 and *search* = 0 (the dataset does not change at the macro level) and sweep parameter spaces for the *threshold* and *backoff* values. Then $w_{max}$ and/or $w_0$ can be adjusted, followed by another parameter sweep, until acceptable accuracies are achieved.

### 5.6  Spike Sorting Results

#### Benchmark

The spike sorting benchmark consists of a rat hippocampal data set from Chung et al. [11]. Filtering and spike detection is done prior to encoding, and simulations begin with the encoding step.

The benchmark consists of 10390 spikes, and spikes were given 6 labels by the benchmark authors, indicating that 6 neurons are represented in the data set. Normally in spike sorting the number of neurons is unknown at the beginning, but our objective in this section is to illustrate the clustering capability of a dendrite, not to tackle the general problem of determining the number of clusters, i.e., the number of neurons being sampled.

One feature of the benchmark is that spikes from different neurons are not evenly distributed in time. Rather, a particular neuron or pair of neurons may be the only ones spiking for extended periods of time (see Figure 16). If anything this makes the problem more challenging than if the neuron spikes are evenly distributed.

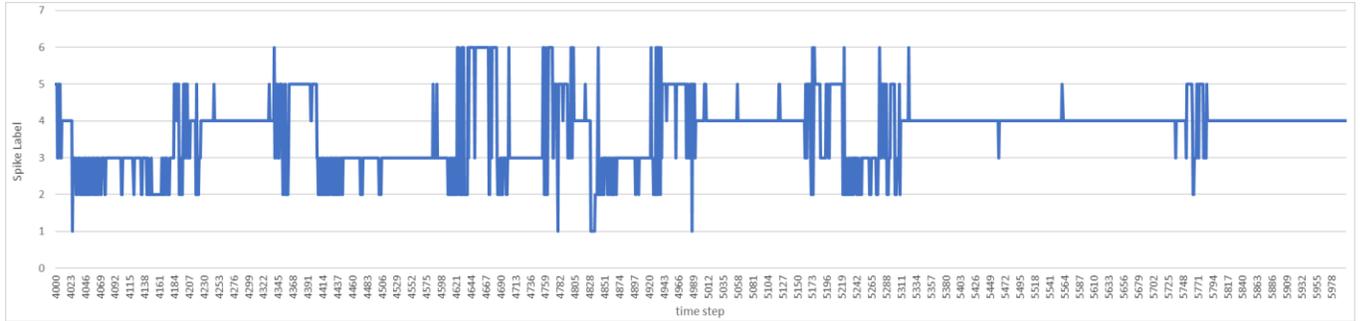

**Figure 16.** A section of the spike sorting benchmark showing the neurons that spike as a function of time. Individual neurons or pairs of neurons dominate extended periods of time.

*Metrics*

Cost is measured as the number of synaptic weights, *num_weights*. Given the frame dimensions, (*before*, *after*, and *precision*), *frame_size* = $2^{precision}$ × (*before* + *after* + 1). Then, given the number of clusters (6 in this example), the synaptic crossbar cost is *num_weights* = *frame_size* × *num_clusters*. In general, we would like to keep frame size parameters small in order to reduce costs, while still giving satisfactory results

The quality of clustering is measured by the average distance, *avg_dist*, over all the input patterns. The lower *avg_dist*, the better the clusters.

An indirect quality measure is *wt_convergence* = $\Sigma w_{ij}*(w_{max} - w_{ij})/(w_{max} \cdot num\_weights)$. The $w_{max}$ in the denominator normalizes the metric. An advantage of the *wt_convergence* metric is that it can be easily computed for sequences of individual time steps. A good set of clusters having a low *avg_dist* typically have weights that converge to a bimodal distribution so *wt_convergence* is close to 0.

*Frontend Optimization*

Using a *k*-means back-end, the parameter space was simulated, and the encoding parameters below were determined to yield a good compromise between cost and performance.

| frame size | | | downsample | similarity coding | | | | metrics | |
|---|---|---|---|---|---|---|---|---|---|
| before | after | precision | stride | block size | row hot | col hot | # centroids | av_dist | cost |
| 6 | 6 | 5 | 3 | 5 | 1 | 3 | 6 | 0.73 | 2496 |

Figure 17a shows the six centroids found by *k*-means operating on integer vectors. The centroids for the pixel version for 3-hot column similarity coding are in Figure 17b.

*SDP Clustering*

Given the encoding parameters given above, SDP was used for clustering. $w_{max}$ = 12 and $w_0$ = 8. The *capture* parameter was set to 1 and the *search* parameter to 0. Then values for the *threshold* and *backoff* were systematically swept yielding the parameter values and metrics given below.

| clustering params | | | | metrics | | |
|---|---|---|---|---|---|---|
| thresh | capt | backoff | srch | wtConv | av_dist | cost |
| 128 | 1 | 9/16 | 0 | 0.008 | 0.71 | 2496 |

Because *capture* was set to 1 in all simulations, *backoff* can be a fraction. For an integer implementation, these are scaled by multiplying by the *backoff* denominator.

Note *avg_dist* is slightly better than with *k*-means. To be fair, *k*-means has lots of local optima depending on initial seeds. For *k*-means simulations, 64 different pseudo-random initial centroids were used. Given this small sample of initial centroids, it is very likely that some other seed will lead to an *avg_dist* that is at least as good as the SDP version, however any difference will likely be small.

*Visualizing Clustering Quality*

Applying conventional *k*-means to numerical vectors (Figure 14b) yields the centroids shown in Figure 17a.

Applying *k*-means to the binarized (pixel) encoding with 3-hot column similarity yields the centroids in Figure 17b in heat map form (green = 0, red = 1).

When SDP learning is used, the final weight matrices for each of the clusters are shown in Figure 17c. Because the SDP method tends to convergence to a bimodal weight distribution, the weights are only an approximation of the centroids.

Finally, when the dataset is applied with the SDP-determined weights, the centroids for the resulting clusters are in Figure 17d. Although not exactly the same as the *k*-means centroids (Figure 17b), they are quite similar.

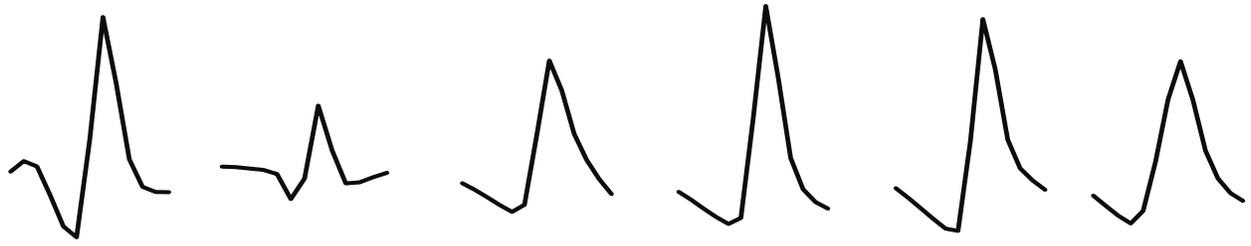

a) integer *k*-means centroids

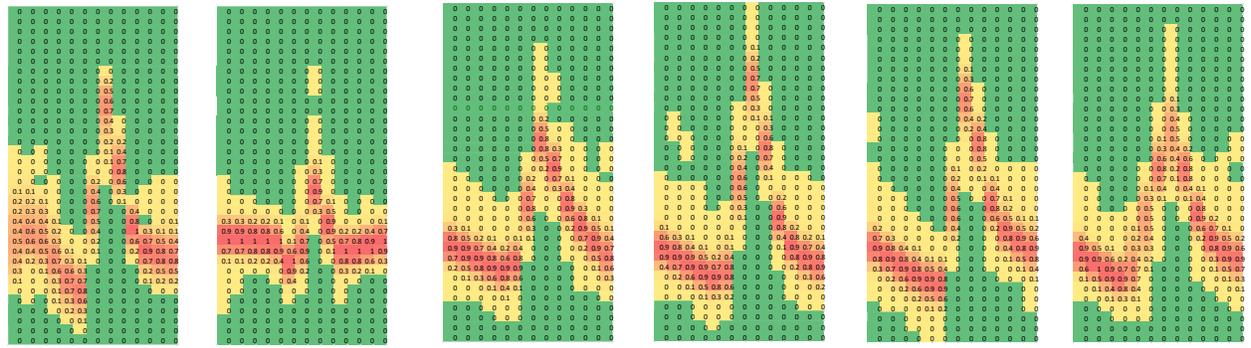

b) 3-hot similarity coded *k*-means centroids

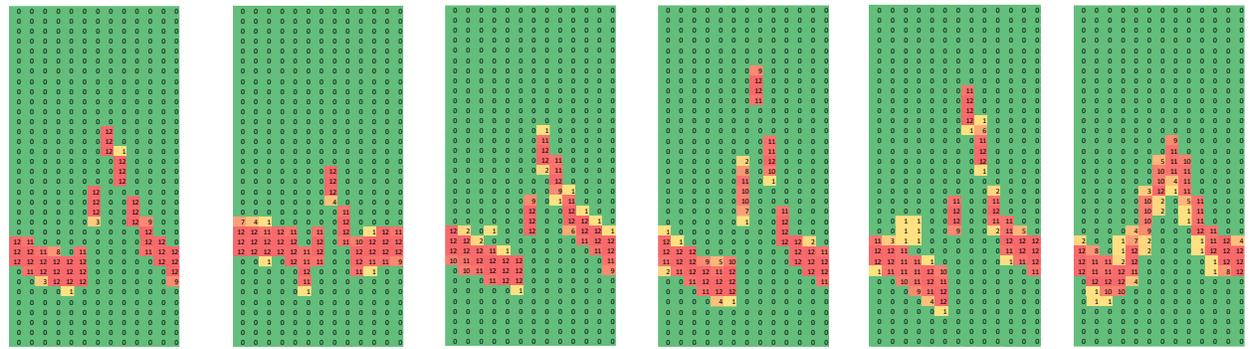

c) SDP weights

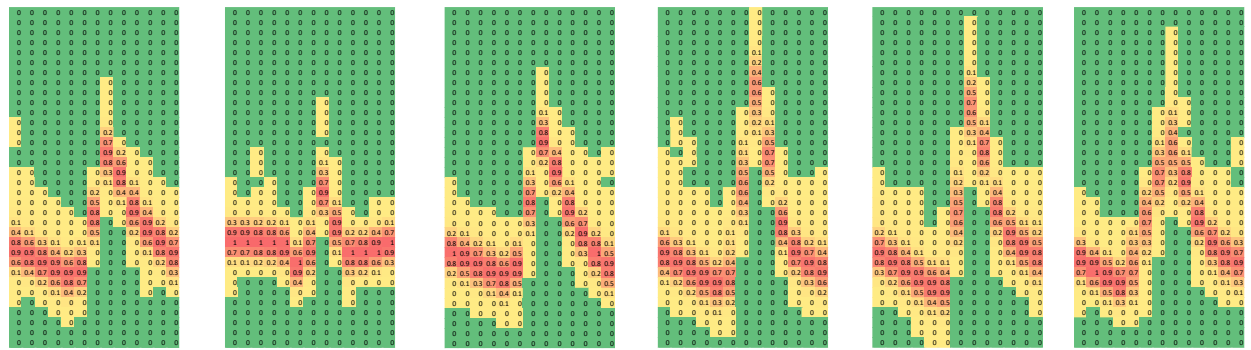

d) SDP centroids

**Figure 17. Results for spike sorting benchmark. a)** *k*-means centroids for inputs given as integer vectors, **b)** the pixel versions with 3-hot similarity encoding in the vertical dimension, **c)** The weights for the 3-hot similarity coded inputs after SDP. **d)** The cluster centroids when input patterns are applied to a dendrite with the weights given in c). Although not exactly the same as the *k*-means centroids in b), they are very similar.

*Online Adaptivity*

One of the key features of online SDP is that centroids dynamically adapt to changes in input patterns. This is demonstrated in the following simulations.

The full benchmark sequence of 10,390 spikes was first applied to the network, then beginning with 10,391, the individual waveforms were flipped top-to-bottom and applied a second time. This is a very immediate and disruptive change in the input stream. Recall that the SDP entry $x_i=1$, $z_j=0$ causes the weight to be incremented: $\Delta w^i_j = +search$. As long as $z_j$ continues to be 0, the weight $w^i_j$ will edge up by $+search$ until it reaches the baseline weight $w_0$.

To determine $w_0$: if the input vector $x$ has $m$ ones, then define $w_0 = threshold/m$. Consequently, as long as $z_j=0$, if a given pattern is applied a sufficient number of times, then all the weights corresponding to the 1s in the input pattern will eventually reach $w_0$. Because there are $m$ of them, their sum will eventually reach the *threshold*, and the given neuron is eligible for WTA inhibition. If no other neuron reaches a higher potential, then the given neuron will "win", and the *capture / backoff* mechanism will begin to solidify the given pattern as a nascent centroid.

Results for two different *search* values are in Figure 18. In these simulations, weight convergence is used as an indirect measure of clustering quality. As before *capture* = 1 and *backoff* = 9/16. When *search* = 1/256, the step where the input patterns are flipped is quite evident. Eventually, after about 5000 time steps, the weights converge to their original levels. With a larger *search* = 25/256, the disruption is just as great; however, the weights reconverge much more quickly. Note that with the higher *search* parameter, the weight convergence value contains more "noise" (for lack of a better term). Informally, this indicates the SDP mechanism is more open to cluster adaptations. This reflects the classic stability/plasticity tradeoff: the greater the plasticity (as reflected in the higher *search* parameter), the less the stability.

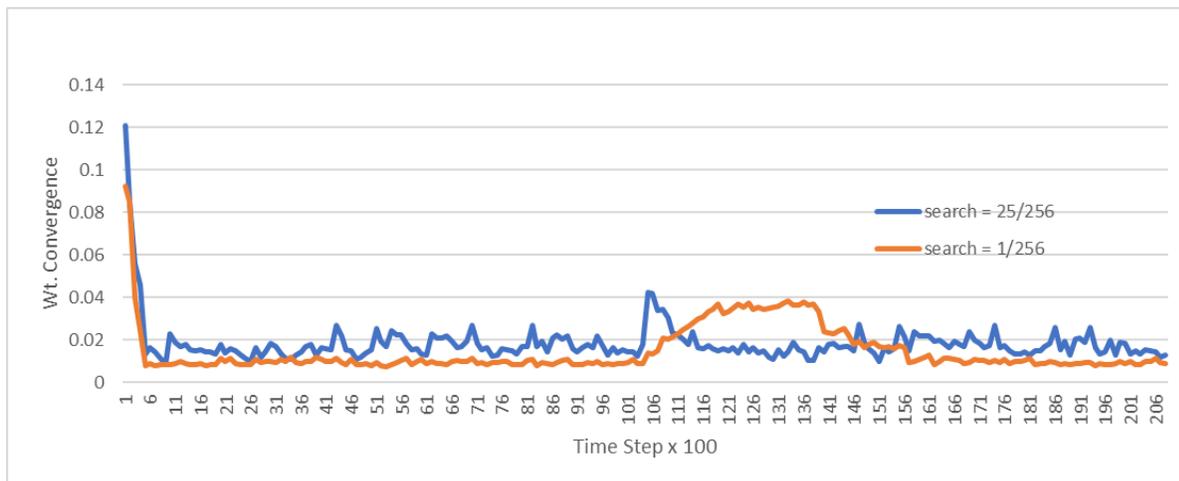

**Figure 18. Weight convergence versus time steps (number of input patterns) for two *search* parameters.**

## 6. Online Supervised Classification with Active Dendrites

An online implementation continually takes a sequence of inputs and produces a sequence of outputs in a streaming fashion. At every step it infers an output based on what it has learned up to that point. Supervised learning may be applied continually or intermittently. When learning is turned on, an output is first inferred, then the correct output is provided for weight updating.

### *6.1 Clustering Voters*

In this section a network composed of active dendrites (essentially single-dendrite neurons) is constructed. For this application, the output of an active dendrite is binarized. See Figure 19. The binarized outputs are interpreted as "votes" which are tallied in order to identify classes. For supervised classification, the modified dendrite combines the function of **c**lustering and **v**oting, so it is referred to as a *CV unit*.

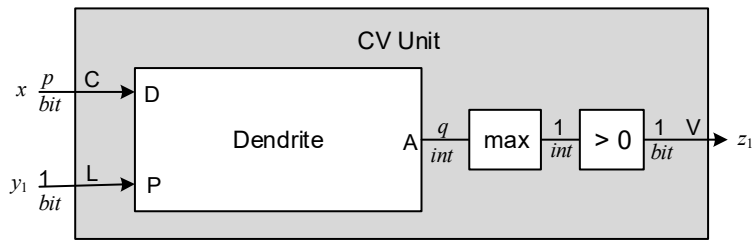

**Figure 19 A clustering voter (CV) unit is an active dendrite with a binarized output. Each CV unit is associated with a specific label. The input C carries a context – a pixel pattern in the case of MNIST, and the input L is active when the CV unit's associated label is present. The output V is a vote for a given label that is fed to a tally unit. In terms of the larger neuromorphic architecture (Figure 2), this is a neuron having a single dendrite.**

For inference all the labels are active, and for learning only a single label is active (provided by a supervisor).

An input image or pattern to be classified is divided into potentially overlapping receptive fields (RFs). Each RF is associated with a CV *group* containing a CV *unit* for each label (Figure 20). Each CV unit updates weights when its associated label is applied, independently of all the other CV units in its group. In effect, each CV unit forms clusters for its associated label. During inference, a CV unit casts a "vote" for its associated label when the input pattern is a good match (based on the dendrite's threshold). Each CV unit's vote is produced independently of all the other CV units, so a given CV group may cast multiple votes (and often does).

### *6.2 Extended Example: Online MNIST*

The architecture is illustrated via the construction of a system that implements an online version of the classic MNIST benchmark [17]. See Figure 20.

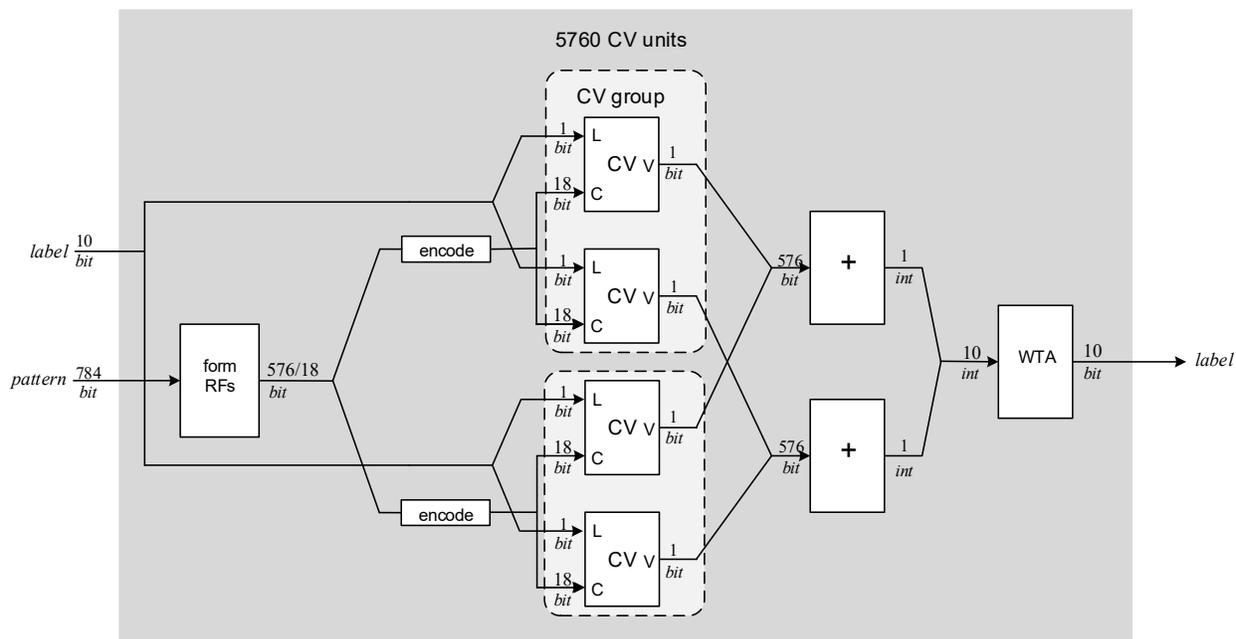

**Figure 20. An online supervised classifier for the online MNIST benchmark. It contains 576 CV groups, one for each RF. Each CV group consists of 10 CV units, one for each label. There is a total of 5760 CV units.**

Receptive field (RF) formation and encoding functions are application dependent. RF formation divides the input into (potentially overlapping) RFs. For the MNIST application, the input patterns are 784 bit (28x28) images of the numerals 0 through 9. See Figure 21. For the benchmark study here, the images are converted to black-and-white. RF formation divides the 28×28 input pattern into all overlapping 5×5 RFs, yielding a total of 576 RFs. In each RF, a checkerboard pattern of pixels forms the actual inputs. In the example, 9 of the 25 pixels are chosen. So, each RF yields is a 9-bit vector.

For MNIST, the input RFs are further encoded to a constant weight code. In this case, a two-rail encoding is used, so each RF is expanded to an 18-bit vector.

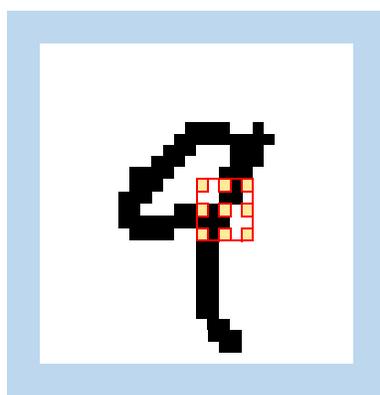

**Figure 21. Example MNIST image with an example 5×5 RF in red, and selected checkerboard pattern of 9 pixels in yellow.**

After encoding, the 18-bit vectors are distributed to 576 CV groups, each group containing 10 CV units – one unit for each of the ten classes.

During supervised learning, a label is supplied, and each of the 10 CV units within a group clusters the input patterns associated with its specific label. Each label essentially has its own private set of clusters. During inference, all the labels are activated, and if the dendrite contained in a CV unit has a good match (its threshold is reached), the resulting spike is interpreted as a binary vote for the CV unit's label.

Then, on the right side of Figure 20 all of the votes for each of the classes are tallied (summed), and the WTA block selects the winning class as output. The one-hot WTA winner identifies the label of the applied input pattern.

*Design Method*

To implement a supervised classifier, the designer determines the following.

1) RF formation and encoding – these functions are both highly application dependent and crucial to the quality (accuracy and cost) of the classifier.

2) Network sizing: The number of RFs, the number of CV units per CV group (the number of labels), and the number of segments per CV unit.

3) Network tuning: For a representative data set, the parameters $w_{max}$, $w_0$, *threshold*, *capture*, *backoff*, and *search* are established via parameter sweeps. During normal operation, these global parameters are fixed and are the same across all the segments in the network. To expedite the process a (small) representative subset of RFs may be used.

*Benchmark results*

MNIST contains 60K training inputs and 10K test inputs. The test inputs have a separate set of writers than the training inputs. For the first benchmark simulations, the training and test inputs are combined to form a length 60K + 10K input stream that is applied only once. Learning begins immediately and continues throughout the run. In Figure 22, the error rate is computed for each block of 1000 inputs. After a few thousand inputs, error rates below .1 are achieved. Then gradually over time until the 60000[th] input, the error rate trends downward with rates approaching .02. After input 60000, the 10000 inputs from new writers are applied. At that point, the error rate ticks up, but then settles back down, reaching down to .0055. The overall error rate for the last 10000 test inputs is .037.

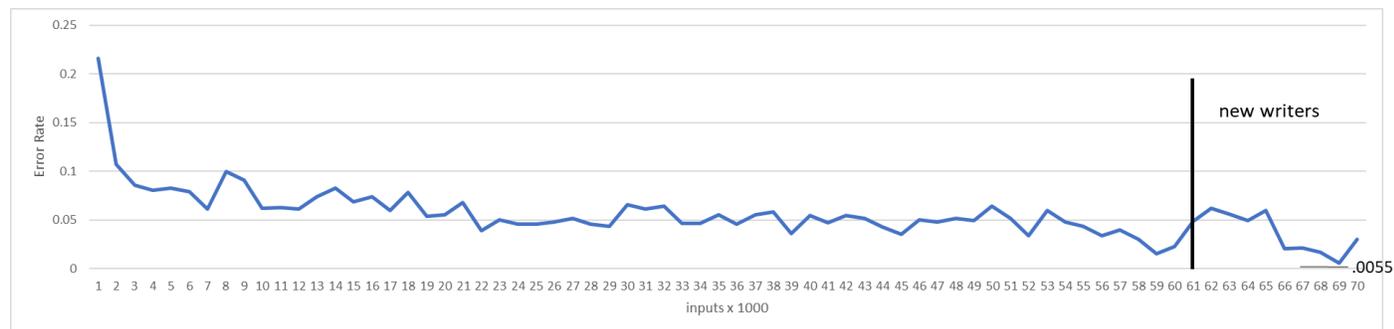

**Figure 22. Online accuracy for single pass through 60K + 10K MNIST benchmark. Error rate is computed for blocks of 1000 inputs.**

Observe that each CV group has a CV unit for each label. These CV units vote independently, so each CV group typically votes for more than one label, several labels in fact. Figure 23 shows the vote totals for the first 20 images in the test set that begins with image 60001. There are 576 CV groups and the vote totals for each label are in the 300 to 500 range. For these 20 examples, on average a CV group votes for 6.9 different labels for each image. This suggests that it is by eliminating less likely labels (the ones not voted for) that the correct label is determined.

|  | 0 | 1 | 2 | 3 | 4 | 5 | 6 | 7 | 8 | 9 | Label |
|---|---|---|---|---|---|---|---|---|---|---|---|
| 60001 | 428 | 380 | 452 | 433 | 434 | 423 | 371 | 545 | 402 | 495 | 7 |
| 60002 | 381 | 323 | 470 | 425 | 298 | 375 | 416 | 289 | 359 | 285 | 2 |
| 60003 | 415 | 549 | 467 | 457 | 476 | 453 | 438 | 472 | 479 | 435 | 1 |
| 60004 | 485 | 297 | 357 | 375 | 330 | 392 | 406 | 355 | 347 | 332 | 0 |
| 60005 | 378 | 349 | 403 | 374 | 524 | 391 | 378 | 439 | 395 | 470 | 4 |
| 60006 | 410 | 554 | 456 | 445 | 481 | 446 | 416 | 481 | 477 | 439 | 1 |
| 60007 | 338 | 354 | 378 | 384 | 465 | 396 | 332 | 396 | 413 | 406 | 4 |
| 60008 | 343 | 390 | 404 | 407 | 419 | 425 | 366 | 405 | 406 | 416 | 9 |
| 60009 | 360 | 318 | 372 | 305 | 372 | 394 | 367 | 333 | 347 | 339 | 5 |
| 60010 | 375 | 326 | 382 | 360 | 444 | 398 | 326 | 478 | 394 | 487 | 9 |
| 60011 | 540 | 266 | 392 | 388 | 326 | 403 | 396 | 309 | 384 | 328 | 0 |
| 60012 | 390 | 281 | 322 | 317 | 334 | 353 | 459 | 302 | 322 | 295 | 6 |
| 60013 | 403 | 365 | 412 | 429 | 501 | 453 | 385 | 483 | 449 | 543 | 9 |
| 60014 | 489 | 265 | 362 | 385 | 353 | 400 | 392 | 358 | 390 | 374 | 0 |
| 60015 | 404 | 540 | 440 | 465 | 449 | 441 | 418 | 441 | 466 | 430 | 1 |
| 60016 | 373 | 290 | 372 | 442 | 318 | 456 | 334 | 330 | 399 | 325 | 5 |
| 60017 | 400 | 333 | 407 | 389 | 471 | 411 | 360 | 471 | 409 | 517 | 9 |
| 60018 | 428 | 347 | 425 | 422 | 392 | 403 | 357 | 531 | 370 | 459 | 7 |
| 60019 | 274 | 242 | 327 | 355 | 281 | 340 | 301 | 274 | 304 | 261 | 3 |
| 60020 | 395 | 407 | 413 | 428 | 517 | 450 | 399 | 456 | 422 | 446 | 4 |

**Figure 23. Vote totals for MNIST images 60001 through 60020. The rows contain the number of votes for each label, and the correct label is the rightmost column. The highest vote total is shaded green. Of the 20 images the correct image gets the highest vote total with one exception, image 60008. In that case the highest total is 425 and the total for the correct label is 416.**

Simulation for different segment counts were done with the error rates given in Figure 24. The costs vary linearly with segment counts, while error rates are sublinear. Doubling cost from 8 to 16 segments reduces error rate by .007. Adding another 8 segments reduces error rate by .004.

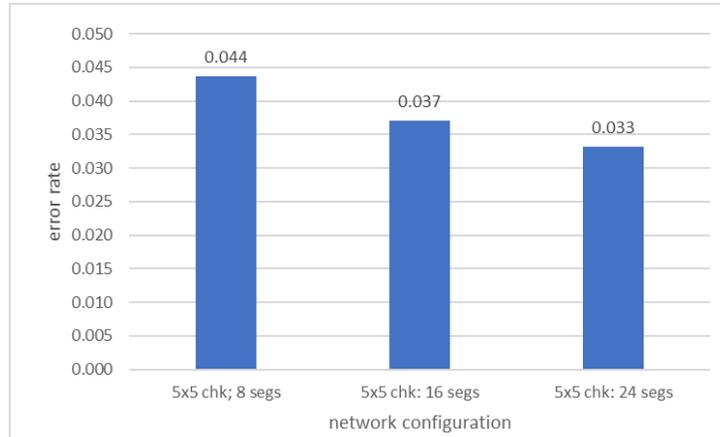

a) Error rate

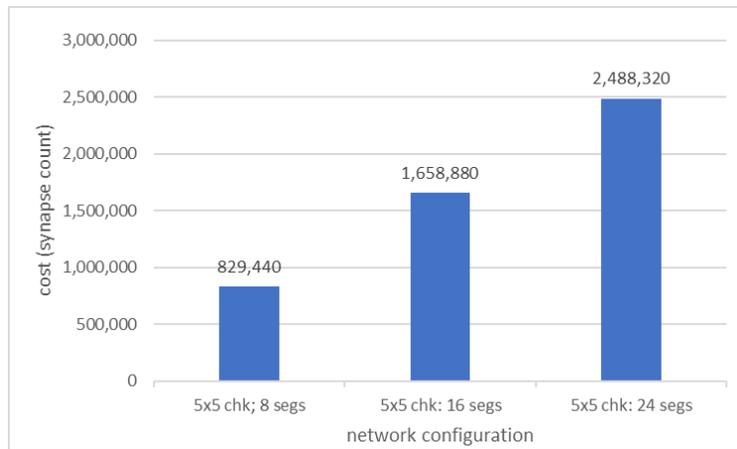

b) Cost as measured by synapse count

**Figure 24. Error rates and costs for networks as segment counts are varied.**

*Adaptivity*

To illustrate adaptivity when inputs change at the macro level, the patterns are transposed after the first 30K inputs. See Figure 25. There is an immediate increase in error rate as expected, but after about 7K inputs, error rates are reduced to the same levels as in the original 60K + 10K simulations.

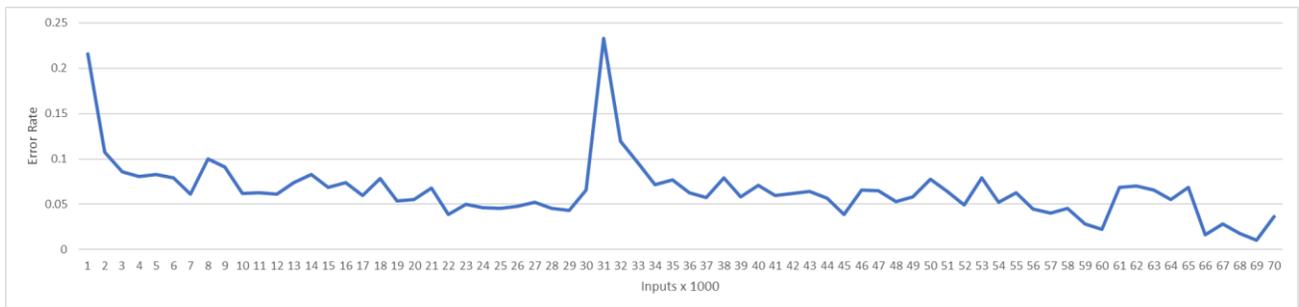

**Figure 25. Error rates per 1000 inputs, where the 60K+10K input stream is transposed after the first 30K inputs.**

## 7. Discussion

The proposed neuromorphic architecture is simple and easy to understand. It starts with basic point neuron functionality, "segments" in the active dendrite model, and combines segments to form dendrites that are capable of clustering input patterns. For learning it uses SDP, a binarized version of the classic STDP. Consequently, learning is localized, and online operation comes naturally. Furthermore, although not explored in this document, recurrent networks require no special handling.

As discussed in Section 1.3, SNNs have been widely studied for many years, so it is natural to ask what sets this model apart.

• An active dendrite model [15], although not original in this work, has yet to be widely adopted. It is significantly more powerful than a point neuron, yet most SNN research continues to use the point neuron model.

• A segment that is akin to the classic point neuron, except it does not fire spikes, rather it emits the value of the dot product. A WTA unit then uses these values to select a winner, and in the process, triggers SDP updates.

• Constant weight codes (or nearly constant weight) that are also sparse. If number of spikes is constant, a surrogate for the centroid distance is the sum of the weights for the spike inputs. If only weights for spike inputs are summed and spike inputs are sparse, then the number of weights to be summed is relatively small, which leads to high efficiency.

• A voting mechanism employing independent CV units that essentially withhold votes for the most unlikely labels, thereby predicting the correct label with the fewest "no" votes.

• Networks for clustering and classification that are only one layer deep, but very wide. This goes against the machine learning trend toward deep networks. It is speculated that as these networks are scaled up, they will grow in width and not in depth.

### *7.1 Seminal Prior Research*

As pointed out in the introduction, many individual elements of the model were first proposed years ago. It is the way the components are combined that sets this model apart.

A *point integrate-and-fire neuron* was proposed by McCulloch and Pitts over eighty decades ago [24]. A similar function is performed by a *segment* in this work, instead of firing a spike when the threshold is reached, a segment passes its value to a WTA network, and the WTA network outputs the segment with the highest value. This mechanism allows segments to interact and provides greater overall functionality than if the segments independently emit binary values. This interaction via WTA is key to implementing a clustering function.

The *dendrite* functionality was anticipated about 25 years ago by Natschläger and Ruf [25] who combined point neurons with a WTA network [27] to implement a clustering function.

The *learning rules* used in the model are more-or-less binarized versions of STDP as studied 25 years ago by Bi and Poo [10], and Gerstner et al. [13].

Much of the prior research that employed point neurons feeding a WTA unit is based on temporal spike encoding [19]. Research by the Thorpe group [22][23] used SNNs with STDP for updates, and arguably reached the high water mark for this type of network.

Most of the author's prior work [1]-[8] followed the Thorpe lead and employed temporal coding. The model used here is essentially a degenerate temporal model where the temporal precision is reduced to 1 bit. Consequently the temporal model becomes a binary model. With a temporal spiking model, the function of generating an output spike is realized as the summation of temporal response functions [14] where weights determine the amplitudes of the functions. With a single bit of temporal precision, this is reduced to summing weights directly.

With *active dendrites* [15], groups of parallel clustering dendrites are controlled by enable signals (proximal inputs) that allow multiple dendrites to act in concert. When active dendrite outputs are combined via a voting mechanism, they implement supervised classification. Labels are mapped to proximal inputs that enable each dendrite to cluster shared context information (distal inputs) associated with a given label.

The centralized method of tallying votes does not appear to be "neuromorphic". As proposed by Hawkins [16], however, connections via apical dendrites can implement a form of distributed voting. Consequently, if the objective is vote tallying, then the centralized method used here is neuromorphic in function although not in structure.

### *7.2 Future Research Directions*

This document is focused on the two lower architecture levels shown in Figure 2. A natural research path is to consider the next higher layers, i.e., continue on a neuromorphic path. A second path is to start with the two lower layers, which are capable of clustering and classification, and use them to build larger non-neuromorphic architectures.

*Neuromorphic Path*

In this work, computation is done at the dendrite level. To form a complete neuron, multiple dendrites are max-ed together. A CV unit is essentially a neuron having a single dendrite. Networks using multi-dendrite neurons are employed when constructing minicolumns and macrocolumns as in [9].

In [9], a macrocolumn is constructed of grid and place cells in a manner that is heavily influenced by Hawkins/Numenta research summarized in [16]. Macrocolumn architecture is demonstrated by simulating a benchmark based on one used by Lewis et al. [18]. The research benchmark involves a mouse navigating 2-dimensional environments in the dark (only features at the current location can be sensed). Because a single macrocolumn can hold multiple environments at the same time, the mouse is introduced to a sequence of environments, say a few tens, and has the opportunity to explore and learn each of them. As initial explorations take place, the macrocolumn learns spatial relationships among features within each of the environments.

Then, the mouse is dropped into an arbitrary environment at an arbitrary location. It first orients itself by moving about in the environment, associating features it senses with what it previously learned. Assuming no inherent ambiguity in the environments, this process eventually converges to a unique feature within a known environment: the mouse is oriented. In the paper, a naive agent is used to illustrate the basic ability to acquire navigational information from the macrocolumn. A more elaborate agent employing reinforcement learning is under development [8].

The importance of this research goes beyond navigating a physical space. It is conjectured that all of the neocortex is based on grid/place cells, and the grid is defined over an abstract conceptual space. Then, if "thinking is a form of movement" [16], navigating through physical space as in [8] may lead to future methods for navigating conceptual space – i.e. true brain-like thinking.

*Non-neuromorphic Path*

If one targets conventional machine learning applications, then the research trajectory of the proposed neural network paradigm shouldn't require a decades long path like the one taken during the development of conventional deep neural networks. A more aggressive approach is to choose a research target that is at the forefront of today's machine learning systems.

Given a few million weights, it is shown in Section 6 that the supervised classification system does an excellent job implementing an online version of the MNIST benchmark. ChatGPT-style LLMs are extremely large, supervised classifiers. Hence, a strong motivation for pursuing LLMs as a research target is that, given a trillion weights, one can construct networks composed of interconnected clustering and classifying subsystems (of huge scale). Observe that these subsystems are focused on finding "similarity", and LLMs exploit text similarities on a very large scale. The structure of such an LLM is put forward as wide open topic for future research.

As a whole, such an LLM system would not be neuromorphic – the brain generates streams of words in an entirely different way. However, the clustering and classifying subsystems are neuromorphic.

## Appendix 1: Schematic Notation

The schematics in this paper employ a concise, unambiguous notation in place of conventional equations. They convey the same information in a form that is easy to visualize.

Schematic networks operate on scalars, vectors, and 2-d arrays. Figure 26 contains operators that operate only on lines, so the number of block inputs and outputs are always the same. All indexing begins at 1. Figure 27 contains more complex operators on single lines, including synchronized buffering.

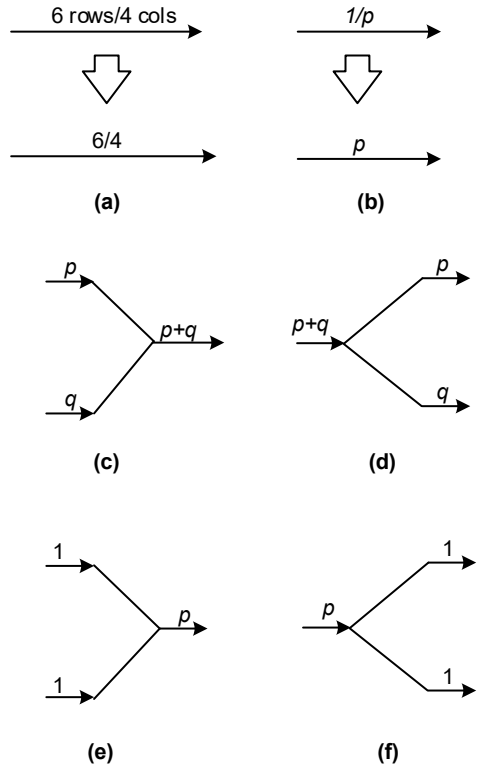

**Figure 26. Schematic blocks that operate on lines. a)** In general, lines carry 2-d arrays, denoted as *#rows / #columns*. **b)** If the number of rows is 1, i.e. 1 / *#columns*, this is simplified to *#columns*. A scalar has one row and one column, 1/1, and is denoted as 1. **c)** a concatenation operator that merges a vector of *p* lines with a vector of *q* lines to form a vector of *p+q* lines. **d)** A vector of *p+q* lines is split to form a vector of *p* lines and a vector of *q* lines. **e)** The special case where *p* scalars are merged to form a size *p* vector. **f)** A size *p* vector is split into *p* scalars (size 1 vectors).

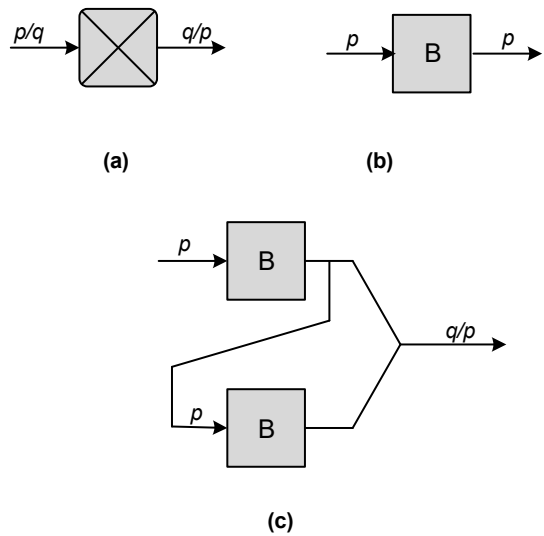

**Figure 27 Schematics for operations on lines. a) the transpose of a 2-d array. b) a single cycle buffer, c) a sequence of *q* cascaded buffers, each buffer holding a vector of size *p*.**

Examples of schematic functional blocks are in Figure 28. Inputs always enter from the left side of a block and exit the right side. The operation to be performed is a function of both the given operator and the dimensions of the inputs and the outputs. All lines leaving a block are labeled with their dimensions, and these labels are an essential part of a schematic. For non-commutative operations, such as division, subtraction, and matrix multiplication, the first operand is the top input to the block.

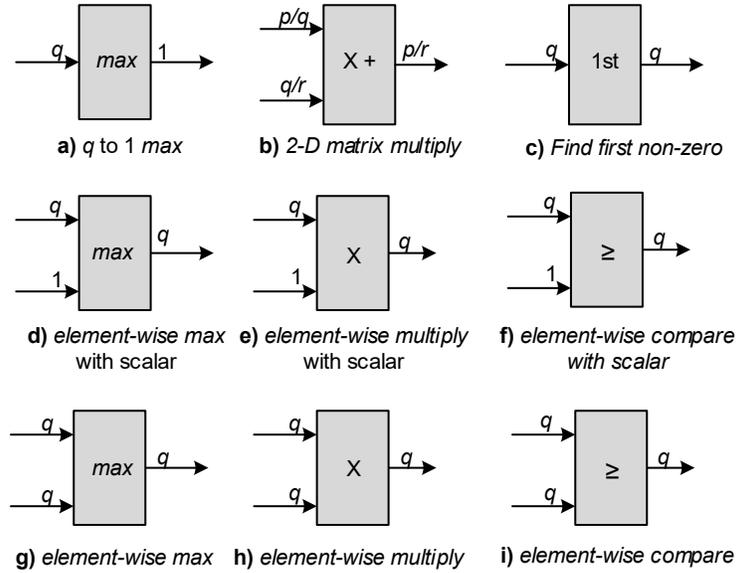

**a)** *q to 1 max*  **b)** *2-D matrix multiply*  **c)** *Find first non-zero*

**d)** *element-wise max with scalar*  **e)** *element-wise multiply with scalar*  **f)** *element-wise compare with scalar*

**g)** *element-wise max*  **h)** *element-wise multiply*  **i)** *element-wise compare*

**Figure 28. Examples of functional blocks.** Three versions of the *max* operation are shown in a), d), and g). The operator notation (*max*) is identical for all 3. However, the numbers of inputs and outputs imply the specific version of *max*. The *max* in a) reduces the $q$ input lines to their maximum value, output as a scalar on a single line. The *max* in d) performs the *max* of a size $q$ input vector and a scalar, producing an output of $q$ lines. Finally, in g) a component-wise vector *max* is performed. A 2-D matrix multiplication consisting of a × followed by + is illustrated in b). The blocks e) and h) are different versions of multiplications with the type of operation being deduced from the input and output line sizes. Operations involving predicates, as in f) and i) always produce bit vectors as outputs. The find *first non-zero operator* c) is essentially a generalized priority encoder. The output is a vector whose $k$th component equals the value of the $k$th of the input vector, where $k$ is the smallest index of a non-zero component in the input. If there is no non-zero component in the input, then the output is set to all zeros. A similar function (not shown) passes a single pseudo-randomly selected non-zero input component $k$ to output component $k$.

Values passed on lines may be either *bits* or integers (*ints*). This is noted below each line as illustrated in Figure 29.

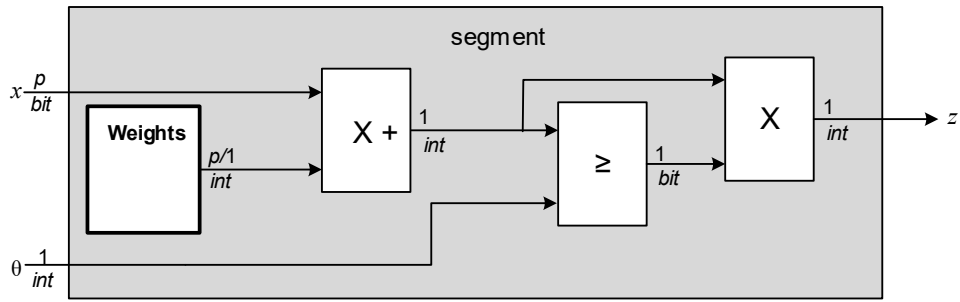

**Figure 29. Schematic example – a segment.** A $1 \times p$ bit vector $x$ is multiplied by a $p \times 1$ integer vector (Weights). The resulting product is an integer scalar. This product is compared with an integer threshold value $\theta$. The result is a single bit which is multiplied by the product to yield the output. The output is an integer that is equal to the product if it is greater than the threshold, otherwise it is 0.